\documentclass{article}

\usepackage{microtype}
\usepackage{graphicx}
\usepackage{booktabs}
\usepackage{amsmath,amssymb,mathtools}
\usepackage{xspace}
\usepackage{pgfplots}
\usepgfplotslibrary{groupplots}
\usepackage{hyperref}
\hypersetup{pdftitle={Finer Parameter Steps for Low-Rank PEFT: A Controlled Study with CP Tensor Adapters}}

\makeatletter
\def\input@path{{icml2026/}}
\makeatother
\usepackage[preprint]{icml2026}

\pgfplotsset{
    compat=1.18,
    every axis/.append style={font=\footnotesize},
    cpcurve/.style={
        mark=*,
        mark size=1.0pt,
        error bars/y dir=both,
        error bars/y explicit
    },
    loracurve/.style={
        mark=square*,
        mark size=1.2pt,
        dashed,
        error bars/y dir=both,
        error bars/y explicit
    },
    confirmmark/.style={
        only marks,
        mark=diamond*,
        mark size=1.8pt,
        mark options={draw=black, fill=white}
    }
}

\newcommand{\R}{\ensuremath{\mathbb{R}}}
\newcommand{\T}{\mathcal{T}}
\newcommand{\tenseed}{\textsuperscript{$\dagger$}}

\icmltitlerunning{Finer Parameter Steps for PEFT}

\begin{document}

\twocolumn[
  \icmltitle{Finer Parameter Steps for Low-Rank PEFT:\\
  A Controlled Study with CP Tensor Adapters}


  \begin{icmlauthorlist}
    \icmlauthor{Xinjue Wang}{1}
    \icmlauthor{Xiuheng Wang}{2}
    \icmlauthor{Yejun Zhang}{3}
    \icmlauthor{Sergiy A. Vorobyov}{1}
    \icmlauthor{Esa Ollila}{1}
    \icmlauthor{Zhi-Yong Wang}{4}
  \end{icmlauthorlist}

  \icmlaffiliation{1}{Department of Information and Communications Engineering, Aalto University, 02150, Espoo, Finland}
  \icmlaffiliation{2}{Universit\'e de Lorraine, CNRS, CRAN, France}
  \icmlaffiliation{3}{Department of Computer Science, Aalto University, 02150, Espoo, Finland}
  \icmlaffiliation{4}{State Key Laboratory of Ocean Sensing, Ocean College, Zhejiang University, Hangzhou 310000, China}

  \icmlcorrespondingauthor{Xinjue Wang}{xjw@ieee.org}

  \icmlkeywords{parameter-efficient fine-tuning, tensor adapters, CP decomposition, LoRA}

  \vskip 0.3in
]

\printAffiliationsAndNotice{}

\begin{abstract}
Low-rank adapters are usually compared by sweeping a small set of ranks, but the rank also fixes the resolution of the parameter budget.
For a $2048{\times}2048$ OPT attention projection, increasing LoRA by one rank stores $4096$ trainable scalars, leaving large gaps between feasible low-budget adapter sizes.
This paper asks whether a tensorized adapter with finer capacity increments changes the observed accuracy--budget trade-off.
We instantiate this question with fixed-component canonical polyadic (CP) tensor adapters.
Under a $32{\times}64{\times}32{\times}64$ tensorization, one normalized CP component stores $193$ trainable scalars per projection, about $21$ times smaller than one LoRA rank step.
We compare CP adapters and LoRA on OPT-1.3B across SST-2, RTE, and BoolQ under matched target modules, training protocol, data caps, and seed schedules.
CP trains stably and fills the gaps between LoRA ranks, but the effect is task-dependent: SST-2 reaches an early low-budget plateau, BoolQ benefits from additional CP components before saturating slightly below LoRA, and RTE remains LoRA-favored.
Finer parameter steps are therefore useful for diagnosing PEFT budget sensitivity, but they do not by themselves guarantee a better accuracy--budget curve.
\end{abstract}

\section{Introduction}

Parameter-efficient fine-tuning (PEFT) adapts a frozen pretrained model by training a small number of additional parameters~\cite{hu2022lora,liu2024dora,Yang2024DecLLMLoraSurvery}.
LoRA is a standard PEFT baseline because its update $\Delta W=BA$ is simple, mergeable, and effective in practice~\cite{hu2022lora}.
The rank $r$ controls both the expressivity and the trainable-parameter count of the update.
This convention is natural, but it also means that the rank fixes the resolution at which the accuracy--budget curve can be observed.
For a large projection matrix, even one rank increment can add thousands of trainable scalars, so the low-budget region may be only sparsely sampled.

Tensorized adapters provide a way to change this budget grid.
After reshaping a matrix update into a multiway tensor, a canonical polyadic (CP) adapter can add capacity one component at a time.
For the OPT-1.3B attention projections studied here, one LoRA rank step stores $4096$ scalars per target projection, whereas one normalized CP component under our main tensorization stores $193$.
This finer grid could reveal useful operating points between LoRA ranks.
At the same time, CP components impose Kronecker-structured directions after reshaping, so finer budget resolution may come with reduced update flexibility.

This trade-off motivates a simple question: when two PEFT families have different parameter-step sizes, should they be compared only at matched budgets, or through the best accuracy each can attain under a budget cap?
We study this question with a best-under-budget comparison: for each budget cap $B$, we report the best tested accuracy from all settings whose trainable-parameter count is at most $B$.
This view makes the discretization of feasible budgets explicit, while avoiding the claim that parameter arithmetic alone predicts accuracy.

\begin{figure}[t]
\centering
\begin{tikzpicture}
\begin{axis}[
    width=0.98\columnwidth,
    height=0.50\columnwidth,
    xlabel={{\sf Stored trainable scalars per target projection}},
    xmin=-80, xmax=4300,
    ymin=0.2, ymax=1.8,
    ytick={0.65,1.35},
    yticklabels={CP, LoRA},
    xtick={0,1000,2000,3000,4000},
    grid=major,
    tick label style={font=\scriptsize},
    label style={font=\scriptsize},
    y dir=reverse
]
\addplot+[only marks, mark=*, mark size=1.4pt]
coordinates {
(0,0.65) (193,0.65) (386,0.65) (579,0.65) (772,0.65) (965,0.65)
(1158,0.65) (1351,0.65) (1544,0.65) (1737,0.65) (1930,0.65)
(2123,0.65) (2316,0.65) (2509,0.65) (2702,0.65) (2895,0.65)
(3088,0.65) (3281,0.65) (3474,0.65) (3667,0.65) (3860,0.65)
(4053,0.65)
};
\addplot+[only marks, mark=square*, mark size=2.2pt] coordinates {(0,1.35) (4096,1.35)};
\draw[dashed] (axis cs:4096,0.35) -- (axis cs:4096,1.55);
\node[anchor=west,font=\scriptsize] at (axis cs:150,0.40) {CP: 193-scalar steps};
\node[anchor=west,font=\scriptsize] at (axis cs:150,1.58) {LoRA: 4096-scalar step};
\node[anchor=center,font=\scriptsize] at (axis cs:2800,1.03) {$\approx 21$ CP steps};
\end{axis}
\end{tikzpicture}
\caption{Discrete budget grids for a single $2048{\times}2048$ target projection. Under the $2048=32\times64$ row/column split, a LoRA rank step stores $4096$ trainable scalars, while a normalized CP component stores $193$.}
\label{fig:budget_grid}
\end{figure}
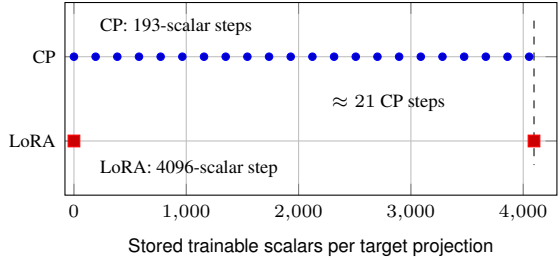

We evaluate this lens on OPT-1.3B with three classification tasks: SST-2, RTE, and BoolQ.
The target modules, trainer, data caps, checkpoint-selection rule, and seed schedules are held fixed across methods.
Matched-budget runs test whether fixed-component CP adapters are viable at LoRA-scale budgets, and a denser CP sweep probes the intervals between LoRA ranks.
Figure~\ref{fig:budget_grid} illustrates the resulting budget grid for one target projection.

The answer is task-dependent rather than uniformly positive.
On SST-2, several very small CP adapters reach an early plateau near $0.93$ accuracy, but no tested CP setting exceeds the best LoRA result under the same budget caps.
On BoolQ, additional CP components matter in the low-budget region before the curve saturates slightly below LoRA.
On RTE, LoRA remains stronger after selected ten-seed confirmation.
These results suggest that finer parameter steps are useful as a diagnostic for PEFT budget sensitivity: they reveal where budget matters, but they do not automatically overcome an expressivity gap.

Our contributions are:
\begin{enumerate}
    \item We make discrete budget steps explicit by comparing PEFT families through the best tested setting under each budget cap.
    \item We instantiate this lens with fixed-component CP tensor adapters and LoRA under a controlled OPT-1.3B protocol.
    \item We show that finer parameter steps reveal task-dependent budget-response patterns: early saturation on SST-2, gradual saturation on BoolQ, and a persistent LoRA advantage on RTE.
\end{enumerate}

We do not claim that the fixed CP adapter studied here is a new state-of-the-art PEFT method.
Instead, we use it as a controlled tensorized adapter family whose capacity parameter has a much smaller step size than LoRA, allowing us to isolate how budget granularity changes the observed accuracy--budget curve.

\section{Related Work}

\paragraph{LoRA variants and rank allocation.}
LoRA~\cite{hu2022lora} established low-rank adaptation as a practical default for PEFT.
Subsequent work modifies the update parameterization or decides how much rank to allocate, including adaptive budget allocation~\cite{zhang2023adaptiveLoRA}, weight-decomposed adaptation~\cite{liu2024dora}, autonomous rank growth~\cite{Sheng2025AprAROMA}, and non-zero initialization~\cite{wang2025nzlora}.
Surveys organize these variants across architecture, optimization, and deployment axes~\cite{Yang2024DecLLMLoraSurvery,Li2026LoRARedux}.
Fixed-rank and Riemannian formulations study complementary geometric aspects of low-rank optimization~\cite{bian2025finding,zhangpilanci2024rplora}.
Our focus is different: we isolate the budget resolution induced by the discrete capacity parameter itself.

\paragraph{Tensorized and structured adapters.}
Tensor decompositions provide classical tools for representing multiway structure~\cite{Kolda2009TensorReview}.
Recent tensorized PEFT methods replace dense low-rank factors with structured tensor parameterizations, including tensor-train formulations such as LoRETTA~\cite{Yang2024LoRETTA}, CP-style adaptations such as LoRTA~\cite{Hounie2024LoRTA} and CaRA~\cite{Veeramacheneni2025CaRA}, and broader structured-update families such as TensLoRA~\cite{Marmoret2025TensLoRA}, AdaZeta~\cite{yang2024Adazeta}, TeRA~\cite{Gu2025TeRA}, and KRAdapter~\cite{Albert2025KRAdapter}.
These methods often combine parameter savings with choices about sharing, tensorization, initialization, or adaptive allocation.
We use a fixed-component CP adapter as a controlled test case rather than as a claim about the best tensor-adapter design.

\paragraph{Budget-aware comparison.}
PEFT methods are often compared at a few matched parameter counts or ranks.
Such comparisons are necessary, but they can hide the fact that different adapter families offer different feasible budget grids.
Our budget-cap view is complementary: it asks what each family can achieve over the tested grid, making budget discretization visible in the evaluation itself.

\section{Comparing Parameter Steps Under Budget Caps}

Let $\mathcal{A}$ be an adapter family with a discrete capacity parameter $k$ and stored trainable-parameter count $P_{\mathcal{A}}(k)$.
We define its \emph{parameter step} as
\begin{equation}
    \Delta P_{\mathcal{A}}(k) = P_{\mathcal{A}}(k+1) - P_{\mathcal{A}}(k).
\end{equation}
For LoRA on a single $m \times n$ target matrix, $P_{\text{LoRA}}(r) = r(m+n)$, so
\begin{equation}
    \Delta P_{\text{LoRA}} = m+n.
\end{equation}
For OPT-1.3B attention projections, $m=n=2048$, giving $\Delta P_{\text{LoRA}} = 4096$ stored trainable scalars per projection.

The parameter step is an accounting quantity, not an accuracy predictor.
It specifies the spacing of feasible budgets for a family.
To compare families under a budget cap, we report the observed best-under-budget curve
\begin{equation}
    U_{\mathcal{A}}(B)
    =
    \max_{k\in\mathcal{K}_{\mathcal{A}}:\,P_{\mathcal{A}}(k)\le B}
    A_{\mathcal{A}}(k),
    \label{eq:budget_curve}
\end{equation}
where $\mathcal{K}_{\mathcal{A}}$ is the tested capacity grid and $A_{\mathcal{A}}(k)$ is held-out evaluation accuracy after best-development checkpoint selection.
Thus $U_{\mathcal{A}}(B)$ describes the best tested setting whose trainable-parameter count does not exceed $B$.

This curve is descriptive rather than a deployment model-selection protocol.
Because CP has more tested capacity values than LoRA, small differences in the best-under-budget curve should not be over-interpreted as reliable wins.
We therefore report the raw curves, matched-budget comparisons, and selected ten-seed confirmations together with the best-under-budget summary.

We report both trainable parameter count and adapter Adam-state memory.
In this setup the two are proportional: one LoRA rank step adds $196{,}608$ trainable parameters across the model and about $1.50$ MB of Adam state, while one CP component step adds $9{,}264$ trainable parameters and about $0.071$ MB.
This memory accounting concerns adapter optimizer state; it is not a measure of total training memory, which also includes the frozen backbone, activations, and implementation-specific overhead.

For LoRA, the tested ranks $\{1,2,3,4,6,8\}$ give six budget levels.
The gaps between them are large: for instance, there is no tested LoRA point between $r=1$ ($196$k parameters) and $r=2$ ($393$k parameters).
A family with smaller parameter steps can place tested settings inside these gaps.
This comparison asks whether those extra settings change the best observed accuracy under the same budget cap.

\section{CP Tensor Adapters}

We use normalized CP tensor adapters as a fixed-component adapter family with smaller parameter steps than LoRA.
All main runs use standard first-order optimization and fix the component count before training.
We call the adapters fixed-component because the value of $c$ is chosen before each run; the experiments then sweep this fixed value across runs.

\paragraph{Adapter form.}
For each target matrix $W_0 \in \R^{m \times n}$, we train
\begin{equation}
    W = W_0 + \Delta W,
\end{equation}
with $W_0$ frozen.
This is adaptation, not compression.

\paragraph{Row/column tensorization.}
We reshape the matrix update by splitting both the row index and the column index.
For the main split, a row index of length $2048$ is written as two indices of sizes $32$ and $64$; the column index is split in the same way.
A $2048{\times}2048$ update is therefore viewed as a four-way tensor,
\begin{equation}
    \T(\Delta W) \in \R^{32 \times 64 \times 32 \times 64}.
\end{equation}
An entry $\Delta W_{a,b}$ is indexed as $\T(\Delta W)_{a_1,a_2,b_1,b_2}$, where $(a_1,a_2)$ identifies the original row and $(b_1,b_2)$ identifies the original column.
This reshape is not optimized; Table~\ref{tab:sensitivity} reports a small sensitivity check over alternative splits.
Each selected projection is tensorized independently, without cross-layer sharing or adaptive component allocation.

\paragraph{Normalized CP parameterization.}
The tensorized update is a sum of $c$ rank-one components with normalized factors:
\begin{equation}
    \T(\Delta W)
    =
    \sum_{s=1}^{c}
    \lambda_s
    u_s^{(1)} \circ u_s^{(2)} \circ u_s^{(3)} \circ u_s^{(4)},
    \qquad
    \|u_s^{(\ell)}\|_2 = 1.
\end{equation}
Each factor $u_s^{(1)} \in \R^{32}$, $u_s^{(2)} \in \R^{64}$, $u_s^{(3)} \in \R^{32}$, and $u_s^{(4)} \in \R^{64}$ has unit norm.
The scalar $\lambda_s$ carries the component scale.
In our implementation, the optimizer stores unconstrained raw factors and normalizes them in the forward pass.
This removes factor-scale ambiguity and was sufficient for stable first-order training.
We count stored trainable scalars rather than the intrinsic degrees of freedom implied by the unit-norm constraints.

\paragraph{Per-component parameter cost.}
One CP component under this tensorization stores $32+64+32+64$ factor entries plus one scalar amplitude, or $193$ trainable scalars per target projection.
By contrast, one LoRA rank step stores $4096$ scalars for the same projection.
After reshaping back to a matrix, a CP component corresponds, up to the index ordering, to
\begin{equation}
    \Delta W_s
    =
    \lambda_s
    \bigl(u_s^{(1)} \otimes u_s^{(2)}\bigr)
    \bigl(u_s^{(3)} \otimes u_s^{(4)}\bigr)^{\!\top}.
\end{equation}
A $c$-component CP adapter therefore gives a matrix update of rank at most $c$, but its rank-one directions are Kronecker-structured rather than arbitrary dense outer products.
The smaller parameter count comes from this restriction.
Across the $48$ adapted projections in OPT-1.3B, one LoRA rank adds $196{,}608$ trainable parameters and about $1.50$ MB of Adam optimizer state.
One CP component adds $9{,}264$ trainable parameters and about $0.071$ MB.
Thus a single LoRA rank step has roughly the same adapter budget as $21$ CP component steps.

\paragraph{Why fix the component count?}
We fix $c$ in advance rather than growing it during training.
This choice isolates parameter-step size from adaptive allocation.
A growing-component rule could be useful, but it would introduce an additional design variable and obscure the budget-resolution question studied here.

\begin{table}[t]
\caption{Stored trainable parameters and Adam optimizer-state footprint over all adapted target projections ($48$ projections: \texttt{q\_proj} and \texttt{v\_proj} in all $24$ OPT decoder layers).}
\label{tab:params}
\vskip 0.05in
\centering
\small
\begin{tabular}{@{}llrrr@{}}
\toprule
Budget & Method & Setting & Params & Opt. MB \\
\midrule
Tiny & LoRA & $r=1$ & 196{,}608 & 1.500 \\
Tiny & CP & $c=21$ & 194{,}544 & 1.484 \\
Low & LoRA & $r=2$ & 393{,}216 & 3.000 \\
Low & CP & $c=43$ & 398{,}352 & 3.039 \\
Mid & LoRA & $r=4$ & 786{,}432 & 6.000 \\
Mid & CP & $c=85$ & 787{,}440 & 6.008 \\
High & LoRA & $r=8$ & 1{,}572{,}864 & 12.000 \\
High & CP & $c=171$ & 1{,}584{,}144 & 12.086 \\
\bottomrule
\end{tabular}
\end{table}

\section{Experiments}

\subsection{Protocol}

All experiments use \texttt{facebook/opt-1.3b}, adapt the \texttt{q\_proj} and \texttt{v\_proj} attention modules (48 target projections), and train with a standard HuggingFace Trainer.
We cap each task at $1000$ training, $500$ development, and $1000$ held-out evaluation examples.
Realized evaluation counts are $872$ for SST-2, $277$ for RTE, and $1000$ for BoolQ.
All runs use $5000$ training steps with evaluation every $1000$ steps, best-development checkpoint selection, and an fp16 backbone.
The base grid uses seeds $0,1,2$; selected confirmation cells use seeds $0$--$9$ under the same protocol.
LoRA uses learning rate $10^{-4}$ and CP uses $2 \times 10^{-4}$, chosen before the main grid.
We did not perform a full per-method learning-rate sweep, so the comparison should be read as a controlled pilot rather than a fully optimized benchmark.

\subsection{Matched-budget comparison}

\textbf{CP trains at LoRA-scale budgets.}
Matched budgets, however, do not imply matched behavior.
Table~\ref{tab:main} reports matched-budget results averaged over seeds $0,1,2$.
CP trains stably in every cell and reaches evaluation accuracy comparable to LoRA on SST-2 and BoolQ.
On RTE, LoRA is consistently stronger.
The matched-budget comparison supports CP viability under this protocol, while also showing that equal parameter counts do not remove method-dependent differences.

\begin{table*}[t]
\caption{Matched-budget results averaged over seeds $0,1,2$. $\Delta$ eval is CP minus LoRA.}
\label{tab:main}
\vskip 0.05in
\centering
\small
\begin{tabular}{@{}llrrrrr@{}}
\toprule
Task & Budget & LoRA dev & LoRA eval & CP dev & CP eval & $\Delta$ eval \\
\midrule
SST2 & Low  & 0.931 & 0.937 & 0.925 & 0.931 & -0.006 \\
SST2 & Mid  & 0.931 & 0.939 & 0.925 & 0.933 & -0.005 \\
SST2 & High & 0.935 & 0.932 & 0.923 & 0.936 & +0.004 \\
RTE  & Low  & 0.768 & 0.747 & 0.770 & 0.732 & -0.016 \\
RTE  & Mid  & 0.789 & 0.753 & 0.762 & 0.722 & -0.031 \\
RTE  & High & 0.771 & 0.745 & 0.760 & 0.729 & -0.016 \\
BoolQ & Low  & 0.761 & 0.741 & 0.744 & 0.735 & -0.006 \\
BoolQ & Mid  & 0.753 & 0.742 & 0.751 & 0.740 & -0.001 \\
BoolQ & High & 0.756 & 0.735 & 0.754 & 0.740 & +0.005 \\
\bottomrule
\end{tabular}
\end{table*}

\subsection{Best under a budget cap}

\begin{figure*}[t]
\centering
\begin{tikzpicture}
\begin{groupplot}[
    group style={group size=3 by 1, horizontal sep=1.15cm},
    width=0.31\textwidth,
    height=0.27\textwidth,
    xlabel={Params (M)},
    xmin=0, xmax=1.65,
    xtick={0,0.5,1.0,1.5},
    grid=major,
    tick label style={font=\scriptsize},
    label style={font=\scriptsize},
    title style={font=\scriptsize},
    legend style={font=\scriptsize, draw=none, fill=none}
]
\nextgroupplot[title=SST2, ylabel={Eval accuracy}, ymin=0.918, ymax=0.950, legend pos=south east]
\addplot+[cpcurve] coordinates {
(0.009264,0.926606) +- (0,0.006976)
(0.018528,0.933869) +- (0,0.009340)
(0.037056,0.929358) +- (0,0.005726)
(0.074112,0.927370) +- (0,0.004774)
(0.148224,0.928517) +- (0,0.003310)
(0.194544,0.930046) +- (0,0.004984)
(0.259392,0.932339) +- (0,0.005734)
(0.333504,0.930810) +- (0,0.007007)
(0.398352,0.931193) +- (0,0.004999)
(0.592896,0.932339) +- (0,0.005255)
(0.787440,0.933486) +- (0,0.006976)
(1.185792,0.934633) +- (0,0.006385)
(1.584144,0.935780) +- (0,0.007520)
};
\addplot+[loracurve] coordinates {
(0.196608,0.937271) +- (0,0.005326)
(0.393216,0.937309) +- (0,0.005297)
(0.589824,0.935780) +- (0,0.011468)
(0.786432,0.938838) +- (0,0.007283)
(1.179648,0.938838) +- (0,0.002886)
(1.572864,0.931957) +- (0,0.005772)
};
\addplot+[confirmmark, forget plot] coordinates {
(0.037056,0.929358)
(0.194544,0.930046)
(0.196608,0.937271)
};
\legend{CP, LoRA}

\nextgroupplot[title=RTE, ymin=0.680, ymax=0.790]
\addplot+[cpcurve] coordinates {
(0.009264,0.712395) +- (0,0.007515)
(0.018528,0.718412) +- (0,0.014440)
(0.037056,0.716005) +- (0,0.023489)
(0.074112,0.726835) +- (0,0.002084)
(0.148224,0.724428) +- (0,0.002084)
(0.194544,0.717208) +- (0,0.007515)
(0.259392,0.737545) +- (0,0.029577)
(0.333504,0.730445) +- (0,0.043070)
(0.398352,0.731649) +- (0,0.019883)
(0.592896,0.735740) +- (0,0.013050)
(0.787440,0.722022) +- (0,0.010830)
(1.185792,0.729242) +- (0,0.016544)
(1.584144,0.729242) +- (0,0.006253)
};
\addplot+[loracurve] coordinates {
(0.196608,0.734055) +- (0,0.025357)
(0.393216,0.747292) +- (0,0.018759)
(0.589824,0.738869) +- (0,0.030704)
(0.786432,0.753309) +- (0,0.019883)
(1.179648,0.759928) +- (0,0.015340)
(1.572864,0.744886) +- (0,0.018526)
};
\addplot+[confirmmark, forget plot] coordinates {
(0.259392,0.737545)
(0.592896,0.735740)
(1.179648,0.759928)
};

\nextgroupplot[title=BoolQ, ymin=0.650, ymax=0.760]
\addplot+[cpcurve] coordinates {
(0.009264,0.662333) +- (0,0.010263)
(0.018528,0.700667) +- (0,0.003055)
(0.037056,0.714333) +- (0,0.007572)
(0.074112,0.721667) +- (0,0.007506)
(0.148224,0.711333) +- (0,0.019502)
(0.194544,0.725667) +- (0,0.013577)
(0.259392,0.730000) +- (0,0.010536)
(0.333504,0.729000) +- (0,0.001732)
(0.398352,0.736500) +- (0,0.011937)
(0.592896,0.738800) +- (0,0.010152)
(0.787440,0.737200) +- (0,0.012372)
(1.185792,0.738667) +- (0,0.006658)
(1.584144,0.739667) +- (0,0.004726)
};
\addplot+[loracurve] coordinates {
(0.196608,0.742700) +- (0,0.012789)
(0.393216,0.741000) +- (0,0.014731)
(0.589824,0.744000) +- (0,0.014933)
(0.786432,0.741667) +- (0,0.005508)
(1.179648,0.736667) +- (0,0.016010)
(1.572864,0.734667) +- (0,0.004933)
};
\addplot+[confirmmark, forget plot] coordinates {
(0.196608,0.742700)
(0.398352,0.736500)
(0.592896,0.738800)
(0.787440,0.737200)
};
\end{groupplot}
\end{tikzpicture}
\caption{Raw accuracy--budget curves with one-standard-deviation error bars over all available seeds. The $x$-axis is trainable parameters in millions; the same ordering also gives adapter Adam-state memory because each trainable parameter has the same optimizer state. Axis ranges are task-specific to show low-budget variation. Most grid points use seeds $0,1,2$; diamond markers indicate selected cells confirmed with seeds $0$--$9$.}
\label{fig:resolution_curve}
\end{figure*}
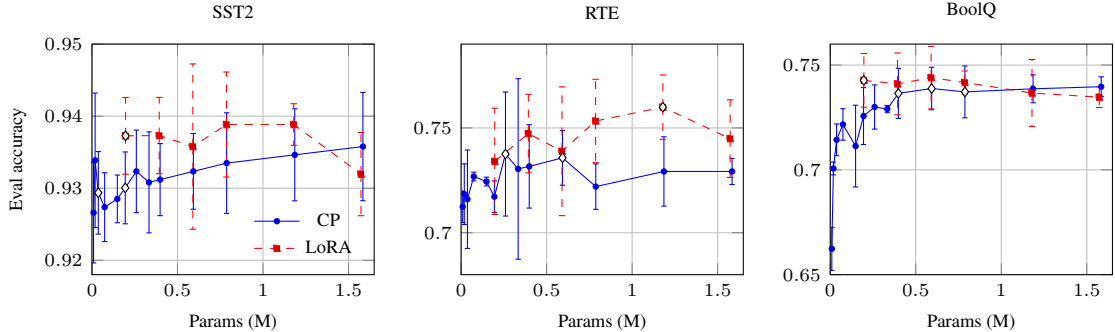

\textbf{Finer CP steps reveal three regimes.}
The main sweep evaluates budgets between the tested LoRA ranks.
We use LoRA ranks $r \in \{1,2,3,4,6,8\}$ and CP component counts $c \in \{1,2,4,8,16,21,28,36,43,64,85,128,171\}$.
Figure~\ref{fig:resolution_curve} shows the raw accuracy--budget curves with one-standard-deviation error bars over all available seeds.
Table~\ref{tab:budget_constrained} reports the descriptive best-under-budget curves defined by Eq.~\eqref{eq:budget_curve}.
The horizontal axis in Figure~\ref{fig:resolution_curve} is trainable parameters, but it can also be read as adapter optimizer-state memory because the two are proportional here.
For example, LoRA $r=1$ and CP $c=21$ both use about $1.5$ MB of Adam state, while the CP points $c=1,2,4,8,16$ all sit before the first LoRA rank.
Selected cells were extended to ten seeds to verify the SST-2 low-budget plateau, BoolQ rise-and-saturation, and RTE persistent-gap patterns suggested by the base grid.

The raw curves are non-monotone, but three task-level patterns are visible.

\textbf{SST-2: early plateau.}
Small CP settings reach accuracy around $0.93$ with a small fraction of the LoRA $r=1$ budget, and more components do not improve the curve monotonically.
The three-seed grid places the best early CP point at $c=2$ ($0.934{\pm}0.009$), but ten-seed checks at $c=4$ and $c=21$ show lower means ($0.929{\pm}0.006$ and $0.930{\pm}0.005$).
We therefore interpret the result as a low-budget plateau rather than as evidence that $c=2$ is a reliable optimum.
No tested CP point exceeds the best LoRA result under the same budget caps on SST-2.
The useful observation is that much of the low-budget region is already close to saturated, and LoRA's sparse rank grid does not resolve that region.

\textbf{BoolQ: rise and saturation.}
The CP curve rises from $0.662$ at $c=1$ to $0.714$ at $c=4$, continues improving to $0.737{\pm}0.012$ at $c=43$, and then saturates near $0.739$ at $c=64$.
The region from $c=1$ to $c=43$ is the informative part of the CP sweep.
Beyond that, adding more components gives little.
The best LoRA result under these budget caps remains slightly higher in mean, with ten-seed LoRA $r=1$ at $0.743{\pm}0.013$.

\textbf{RTE: persistent gap.}
After selected ten-seed confirmation, CP remains below the best LoRA result under the same budget caps.
The confirmed CP means are $0.738{\pm}0.030$ at $c=28$ and $0.736{\pm}0.013$ at $c=64$, while LoRA $r=6$ reaches $0.760{\pm}0.015$.
Under this fixed tensorization and protocol, finer CP steps and additional components do not close the gap.
The higher CP mean at $c=28$ also comes with larger variance, so we do not read it as a precise ordering between the two CP settings.

\begin{table*}[t]
\caption{Best tested accuracy under each LoRA budget. For $B=P_{\mathrm{LoRA}}(r)$, $r^\star$ and $c^\star$ denote the best tested LoRA and CP settings not exceeding $B$. CP budget fraction is the CP parameter count divided by $B$; gap is $U_{CP}(B)-U_{\mathrm{LoRA}}(B)$. \tenseed{} entries use seeds $0$--$9$; all other entries use seeds $0,1,2$. The comparison is descriptive over the tested grid.}
\label{tab:budget_constrained}
\vskip 0.05in
\centering
\scriptsize
\begin{tabular}{@{}llclclcr@{}}
\toprule
Task & Budget & $r^\star$ & $U_{\mathrm{LoRA}}(B)$ & $c^\star$ & $U_{CP}(B)$ & CP budget fraction & Gap \\
\midrule
SST2 & $r=1$ & $1$ & $0.937{\pm}0.005$\tenseed & $c=2$   & $0.934{\pm}0.009$ & 9.4\%  & -0.003 \\
SST2 & $r=2$ & $2$ & $0.937{\pm}0.005$          & $c=2$   & $0.934{\pm}0.009$ & 4.7\%  & -0.003 \\
SST2 & $r=3$ & $2$ & $0.937{\pm}0.005$          & $c=2$   & $0.934{\pm}0.009$ & 3.1\%  & -0.003 \\
SST2 & $r=4$ & $4$ & $0.939{\pm}0.007$          & $c=2$   & $0.934{\pm}0.009$ & 2.4\%  & -0.005 \\
SST2 & $r=6$ & $4$ & $0.939{\pm}0.007$          & $c=2$   & $0.934{\pm}0.009$ & 1.6\%  & -0.005 \\
SST2 & $r=8$ & $4$ & $0.939{\pm}0.007$          & $c=128$ & $0.935{\pm}0.006$ & 75.4\% & -0.004 \\
\midrule
RTE & $r=1$ & $1$ & $0.734{\pm}0.025$          & $c=8$  & $0.727{\pm}0.002$          & 37.7\% & -0.007 \\
RTE & $r=2$ & $2$ & $0.747{\pm}0.019$          & $c=28$ & $0.738{\pm}0.030$\tenseed & 66.0\% & -0.010 \\
RTE & $r=3$ & $2$ & $0.747{\pm}0.019$          & $c=28$ & $0.738{\pm}0.030$\tenseed & 44.0\% & -0.010 \\
RTE & $r=4$ & $4$ & $0.753{\pm}0.020$          & $c=28$ & $0.738{\pm}0.030$\tenseed & 33.0\% & -0.016 \\
RTE & $r=6$ & $6$ & $0.760{\pm}0.015$\tenseed & $c=28$ & $0.738{\pm}0.030$\tenseed & 22.0\% & -0.022 \\
RTE & $r=8$ & $6$ & $0.760{\pm}0.015$\tenseed & $c=28$ & $0.738{\pm}0.030$\tenseed & 16.5\% & -0.022 \\
\midrule
BoolQ & $r=1$ & $1$ & $0.743{\pm}0.013$\tenseed & $c=21$ & $0.726{\pm}0.014$          & 99.0\% & -0.017 \\
BoolQ & $r=2$ & $1$ & $0.743{\pm}0.013$\tenseed & $c=28$ & $0.730{\pm}0.011$          & 66.0\% & -0.013 \\
BoolQ & $r=3$ & $3$ & $0.744{\pm}0.015$          & $c=43$ & $0.737{\pm}0.012$\tenseed & 67.5\% & -0.007 \\
BoolQ & $r=4$ & $3$ & $0.744{\pm}0.015$          & $c=64$ & $0.739{\pm}0.010$\tenseed & 75.4\% & -0.005 \\
BoolQ & $r=6$ & $3$ & $0.744{\pm}0.015$          & $c=64$ & $0.739{\pm}0.010$\tenseed & 50.3\% & -0.005 \\
BoolQ & $r=8$ & $3$ & $0.744{\pm}0.015$          & $c=64$ & $0.739{\pm}0.010$\tenseed & 37.7\% & -0.005 \\
\bottomrule
\end{tabular}
\end{table*}

Table~\ref{tab:ten_seed} reports the selected ten-seed confirmations.
These cells verify the qualitative SST-2 plateau, BoolQ rise-and-saturation, and RTE persistent-gap patterns suggested by the base grid.

\begin{table}[t]
\caption{Selected ten-seed confirmation cells. These runs use the same protocol as the base grid and extend selected settings to seeds $0$--$9$.}
\label{tab:ten_seed}
\vskip 0.05in
\centering
\scriptsize
\begin{tabular}{@{}llrrr@{}}
\toprule
Task & Setting & Params & Opt. MB & Eval \\
\midrule
SST2 & LoRA $r=1$ & 196{,}608 & 1.500 & $0.937{\pm}0.005$ \\
SST2 & CP $c=4$ & 37{,}056 & 0.283 & $0.929{\pm}0.006$ \\
SST2 & CP $c=21$ & 194{,}544 & 1.484 & $0.930{\pm}0.005$ \\
RTE & LoRA $r=6$ & 1{,}179{,}648 & 9.000 & $0.760{\pm}0.015$ \\
RTE & CP $c=28$ & 259{,}392 & 1.979 & $0.738{\pm}0.030$ \\
RTE & CP $c=64$ & 592{,}896 & 4.523 & $0.736{\pm}0.013$ \\
BoolQ & LoRA $r=1$ & 196{,}608 & 1.500 & $0.743{\pm}0.013$ \\
BoolQ & CP $c=43$ & 398{,}352 & 3.039 & $0.737{\pm}0.012$ \\
BoolQ & CP $c=64$ & 592{,}896 & 4.523 & $0.739{\pm}0.010$ \\
BoolQ & CP $c=85$ & 787{,}440 & 6.008 & $0.737{\pm}0.012$ \\
\bottomrule
\end{tabular}
\end{table}

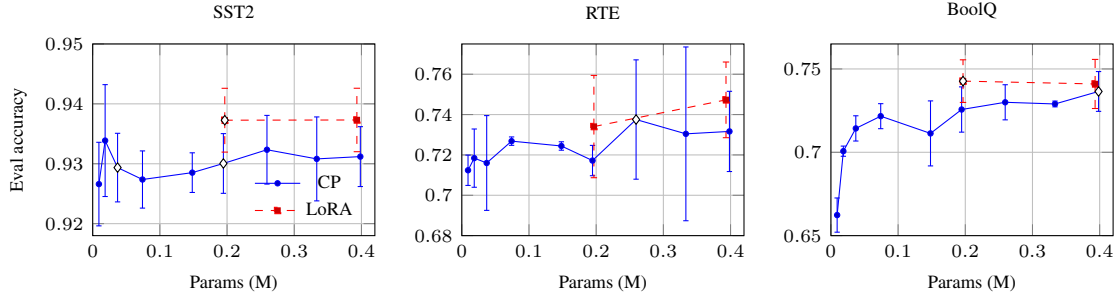
\begin{figure*}[t]
\centering
\begin{tikzpicture}
\begin{groupplot}[
    group style={group size=3 by 1, horizontal sep=1.15cm},
    width=0.31\textwidth,
    height=0.24\textwidth,
    xlabel={Params (M)},
    xmin=0, xmax=0.42,
    xtick={0,0.1,0.2,0.3,0.4},
    grid=major,
    tick label style={font=\scriptsize},
    label style={font=\scriptsize},
    title style={font=\scriptsize},
    legend style={font=\scriptsize, draw=none, fill=none}
]
\nextgroupplot[title=SST2, ylabel={Eval accuracy}, ymin=0.918, ymax=0.950, legend pos=south east]
\addplot+[cpcurve] coordinates {
(0.009264,0.926606) +- (0,0.006976)
(0.018528,0.933869) +- (0,0.009340)
(0.037056,0.929358) +- (0,0.005726)
(0.074112,0.927370) +- (0,0.004774)
(0.148224,0.928517) +- (0,0.003310)
(0.194544,0.930046) +- (0,0.004984)
(0.259392,0.932339) +- (0,0.005734)
(0.333504,0.930810) +- (0,0.007007)
(0.398352,0.931193) +- (0,0.004999)
};
\addplot+[loracurve] coordinates {
(0.196608,0.937271) +- (0,0.005326)
(0.393216,0.937309) +- (0,0.005297)
};
\addplot+[confirmmark, forget plot] coordinates {
(0.037056,0.929358)
(0.194544,0.930046)
(0.196608,0.937271)
};
\legend{CP, LoRA}

\nextgroupplot[title=RTE, ymin=0.680, ymax=0.775]
\addplot+[cpcurve] coordinates {
(0.009264,0.712395) +- (0,0.007515)
(0.018528,0.718412) +- (0,0.014440)
(0.037056,0.716005) +- (0,0.023489)
(0.074112,0.726835) +- (0,0.002084)
(0.148224,0.724428) +- (0,0.002084)
(0.194544,0.717208) +- (0,0.007515)
(0.259392,0.737545) +- (0,0.029577)
(0.333504,0.730445) +- (0,0.043070)
(0.398352,0.731649) +- (0,0.019883)
};
\addplot+[confirmmark, forget plot] coordinates {(0.259392,0.737545)};
\addplot+[loracurve] coordinates {
(0.196608,0.734055) +- (0,0.025357)
(0.393216,0.747292) +- (0,0.018759)
};

\nextgroupplot[title=BoolQ, ymin=0.650, ymax=0.765]
\addplot+[cpcurve] coordinates {
(0.009264,0.662333) +- (0,0.010263)
(0.018528,0.700667) +- (0,0.003055)
(0.037056,0.714333) +- (0,0.007572)
(0.074112,0.721667) +- (0,0.007506)
(0.148224,0.711333) +- (0,0.019502)
(0.194544,0.725667) +- (0,0.013577)
(0.259392,0.730000) +- (0,0.010536)
(0.333504,0.729000) +- (0,0.001732)
(0.398352,0.736500) +- (0,0.011937)
};
\addplot+[loracurve] coordinates {
(0.196608,0.742700) +- (0,0.012789)
(0.393216,0.741000) +- (0,0.014731)
};
\addplot+[confirmmark, forget plot] coordinates {
(0.196608,0.742700)
(0.398352,0.736500)
};
\end{groupplot}
\end{tikzpicture}
\caption{Low-budget zoom ($c \le 43$, LoRA $r=1,2$). Error bars use all available seeds; diamond markers denote selected ten-seed cells. LoRA points correspond to rank choices, while CP points correspond to component counts.}
\label{fig:low_budget_zoom}
\end{figure*}

Figure~\ref{fig:low_budget_zoom} zooms into the low-budget region ($c \le 43$, LoRA $r=1,2$).
This isolates the budget range before and around the first two LoRA ranks.
The first LoRA point is $r=1$, while CP has multiple tested component counts before and around the same memory level.
SST-2 already has several CP points around $0.93$ inside the first LoRA rank interval.
BoolQ rises more gradually and remains below LoRA $r=1$.
RTE has higher variance and no clear low-budget improvement trend.

\subsection{Tensorization sensitivity}

\textbf{The tested reshapes give similar SST-2 results.}
The CP step size depends on how the row and column indices are factorized.
Table~\ref{tab:sensitivity} compares three tensorizations at the SST-2 mid-budget level, with CP component counts adjusted to approximately match the trainable-parameter budget.

\begin{table}[t]
\caption{SST-2 mid-budget tensorization sensitivity, approximately matched trainable-parameter budgets, averaged over seeds $0,1,2$.}
\label{tab:sensitivity}
\vskip 0.05in
\centering
\scriptsize
\begin{tabular}{@{}lrrrr@{}}
\toprule
Split & Dev & Eval & Params & Opt. MB \\
\midrule
row/col $32\times64$ & 0.925 & 0.933 & 787{,}440 & 6.008 \\
row/col $16\times128$ & 0.929 & 0.932 & 790{,}704 & 6.033 \\
row/col $16\times16\times8$ & 0.925 & 0.936 & 789{,}264 & 6.022 \\
\bottomrule
\end{tabular}
\end{table}

All three tensorizations train stably and remain within a narrow evaluation band.
The $16 \times 16 \times 8$ split is strongest in this small study, but the spread is modest.
The reshape affects both step size and component structure, yet the tested alternatives do not indicate brittleness on SST-2.
We therefore treat the $32 \times 64$ split as a representative setting rather than an optimized tensorization.

\section{Discussion}

\subsection{Finer steps as a diagnostic}

Finer parameter steps are most useful when the accuracy--budget curve changes inside a LoRA rank interval.
BoolQ is the clearest example: the CP sweep reveals a gradual rise from very small adapters to the mid-budget region before the curve saturates.
SST-2 shows the opposite pattern, with several low-budget CP points already near the task's plateau under this protocol.
RTE suggests that budget granularity is not the main bottleneck.
Thus the value of finer steps is diagnostic: they reveal the shape of the budget-response curve, even when they do not improve the best-under-budget result.

\subsection{Granularity versus expressivity}

The experiments separate budget granularity from update expressivity.
CP provides many more feasible budgets than LoRA in the same parameter range, but each CP rank-one direction is Kronecker-structured after tensorization.
LoRA rank increments are coarser, yet each increment adds an unconstrained dense rank-one direction.
The RTE gap suggests that, for some tasks, this flexibility matters more than a finer budget grid.
The BoolQ curve suggests the complementary case: a finer grid can reveal useful intermediate capacity growth before saturation, even if the final mean remains slightly below LoRA.

\subsection{Semantically grounded tensorization}

The tensorization used here is imposed on language-model matrices rather than derived from semantic tensor modes.
Domains with native multiway structure may provide a more direct testbed for tensorized adapters.
Wireless communication is one example, where signal and channel objects are naturally indexed by physical axes such as space, time, and frequency, and tensor decompositions are already used for wireless signal processing and channel modeling~\cite{Chen2021TensorWireless}.
In such settings, tensor factors may align with meaningful axes rather than with an arbitrary reshape of a dense matrix.

\subsection{Limitations}

The scope is narrow by design: one model scale (OPT-1.3B), three classification tasks, attention \texttt{q\_proj}/\texttt{v\_proj} targets only, and one main tensorization.
The base grid uses three seeds, with selected key cells confirmed using ten seeds.
Learning rates were chosen before the main grid, without a large per-method sweep.
The tensorization sensitivity study was run on SST-2 only.
Because CP has more tested capacity settings than LoRA, best-under-budget comparisons are descriptive and should be interpreted together with the raw curves and seed confirmations.
The reported memory numbers count adapter optimizer state rather than total training memory.
We also did not grow $c$ during training, tune tensorizations per layer, or allocate different component counts to different projections.
Those choices require separate experiments.

\section{Conclusion}

We studied parameter-step size as a property of PEFT adapter families.
Fixed-component CP tensor adapters provide a much finer budget grid than LoRA under the same OPT-1.3B target projections.
Across SST-2, RTE, and BoolQ, the observed curves are task-dependent: SST-2 shows an early plateau, BoolQ rises and saturates slightly below LoRA, and RTE remains LoRA-favored.
Finer parameter steps therefore help diagnose where the PEFT budget curve changes, but finer granularity alone does not guarantee a better accuracy--budget curve.

\section*{Acknowledgements}
The work was supported by the Research Council of Finland under Grant 359848.

\bibliographystyle{icml2026/icml2026}
\bibliography{reference}

\end{document}